# Object–oriented semantics of English in natural language understanding system

Yuriy Ostapov

*Institute of Cybernetics of NAS of Ukraine, pr. Acad. Glushkova, 40, Kiev, 03680, Ukraine. E-mail: yugo.ost@gmail.com*

**Abstract**

A new approach to the problem of natural language understanding is proposed. The knowledge domain under consideration is the social behavior of people. English sentences are translated into set of predicates of a semantic database, which describe persons, occupations, organizations, projects, actions, events, messages, machines, things, animals, location and time of actions, relations between objects, thoughts, cause-and-effect relations, abstract objects. There is a knowledge base containing the description of semantics of objects (functions and structure), actions (motives and causes), and operations.

## 1. Object–oriented semantics

According to analytic philosophy the world consists of *facts* [5]. Each fact in essence is that some *object* of the real world influences the environment or gets an external effect and changes its state. Facts are interpreted by human being in the form of sentences, which are built from words by means of grammar rules.

Computer understanding of English can be determined as a capability of a program system to translate English sentences into an internal representation so that this system generates adequate (i.e., valid) answers for the questions to be asked by a researcher.

To understand natural language the computer program must have such properties [10]:

- to have *dictionaries* which consist of words of the given language;

- to use the *grammar rules* of this language;

- to contain the description of *knowledge domain*;

- to translate natural language sentences into *an internal representation*;

- to answer questions using *inference algorithms*;

The internal representation of sentences has the dominant role in the understanding system. As the key concept of approach under review is *object*, this approach is named *object-oriented*.

The object acts or changes its state saving integrity and functional identity. The properties of an object are estimated by means of *variables* taking values from appropriate sets. The use of variables enables to describe real objects as predicates of type *person, organization, animal, machine, thing*. The actions of these objects are described by predicates *actions, message, intelligence, job*. The predicate *event* is used for the description of a change of state.

English understanding system, which we called LEIBNIZ, implements mapping of English sentence into database predicates and building inquiry answers. LEIBNIZ includes the database of dictionaries, the database of facts and objects (the **semantic database**), and the database of description of words semantics (the **knowledge base**).

LEIBNIZ works under the control of Windows 2000/XP/7 on IBM PC. VISUAL PROLOG is used as a programming tool [9]. LEIBNIZ is a experimental system and therefore only around 2 400 words are saved in the dictionaries. Volume of the database of facts and objects is determined with the number of sentences of story to be entered by researcher.

Having regard to the knowledge domain, over the course of the last thirty years similar understanding systems have been created by researchers of Stanford and Yale Universities [4, 7, 8]. The main distinction of these works from ours is in the ways of representation of linguistic information and in the use of programming tool for understanding algorithms. In these works the Conceptual Dependency forms (CD forms), the Memory Organization Packets (MOPs), and other data structures are used. By means of these structures different conclusions implementing the problem of understanding are formed. Syntax and semantic properties of words are realized by appropriate program code. This will cause a lot of difficulties in moving to real volumes of used dictionaries and input data which are needed for the solution of practical problems.

As it seems to us, the proposed approach is perspective because it enables to bring about a further advance on the way of creating understanding systems. This advance will be achieved due to providing a powerful *inference engine* and due to *separation* of data representation from program code. The use of inference engine simplifies essentially building inference algorithms. Saving linguistic information in the semantic database provides the effective search of facts and objects to be required. The proposed algorithms will permit to pass on to real volumes of dictionaries and input data. Furthemore, these algirithms can be adapted with comparative ease to another European languages.

## 2. Formal grammar of English

Formal grammar of English can be described by means of the **Backus-Naur form** [1]. For this purpose the conceptual description of the modern English grammar is used [2].

*2.1. The structure of declarative sentence*

⟨declarative sentence⟩ ::= ⟨complex declarative sentence⟩ | ⟨simple declarative sentence⟩

⟨complex declarative sentence⟩ ::= ⟨simple declarative sentence⟩ [⟨coordinating conjunction⟩] ⟨simple declarative sentence⟩

⟨simple declarative sentence⟩ ::= [⟨adverbial modifier⟩] ⟨the group of subject⟩ ⟨the group of predicate⟩ | ⟨the sentence with reverse order of words⟩

The square brackets are used to point to possibility of absence of this element [1].

*2.2. The sentence with reverse order of words*

⟨the sentence with reverse order of words⟩ ::= ⟨the construction of '*there is*'⟩ | ⟨the construction with *there* or *here* at the beginning of sentence⟩ | ⟨the construction with the adverb at the beginning of sentence⟩

⟨the construction of '*there is*'⟩ ::= *there* ⟨*to be* in indefinite tense⟩ ⟨the group of subject⟩ ⟨the group of adverbial modifiers⟩

⟨the construction with *there* or *here* at the beginning of sentence⟩ ::= ⟨*there* or *here*⟩ ⟨the verb of existence in indefinite tense⟩ ⟨the group of subject⟩ ⟨the group of adverbial modifiers⟩
⟨the construction with the adverb at the beginning of sentence⟩ ::= ⟨adverb⟩ ⟨predicate⟩ ⟨the group of subject⟩ ⟨the group controlled with predicate⟩

In the last construction adverbs at the beginning of sentence are *hardly, scarcely, no sooner, never, nothing, not only*, and others.

*2.3. The group of subject*
⟨the group of subject⟩ ::= ⟨basic noun phrase⟩ [⟨determinative construction⟩] | ⟨subject clause⟩
⟨determinative construction⟩ ::= ⟨participial phrase⟩ | ⟨attributive clause⟩ | ⟨infinitive phrase⟩

The basic noun phrase is the group with a noun (for instance, *a little house*) or consists of the groups of noun connected with prepositions: *a little house on the bank of the big river*. The basic noun phrase is conceptually a single whole.

*2.4. The group of predicate*
⟨the group of predicate⟩ ::= ⟨predicate⟩ [⟨the construction controlled with predicate⟩]
⟨predicate⟩ ::= ⟨simple predicate⟩ | ⟨compound verbal predicate⟩ | ⟨compound name predicate⟩
⟨the construction controlled with predicate⟩ ::= [⟨the group of objects⟩] [⟨the group of adverbial modifiers⟩]

*2.5. The group of objects*
⟨the group of objects⟩ ::= [⟨indirect object⟩] [⟨direct object⟩] [⟨prepositional object⟩]
⟨indirect object⟩ ::= ⟨basic noun phrase⟩ [⟨determinative construction⟩]
⟨direct object⟩ ::= ⟨basic noun phrase⟩ [⟨determinative construction⟩] | ⟨infinitive phrase⟩ | ⟨object clause⟩
⟨prepositional object⟩ ::= ⟨preposition⟩ ⟨basic noun phrase⟩ [⟨determinative construction⟩]

The direct object describes the object of influence (for transitive verbs). The indirect object points to the object that a given action is addressed to.

The prepositional object means the object of influence, the subject of action for passive voice, and the way of action.

*2.6. The group of adverbial modifiers*

⟨the group of adverbial modifiers⟩ ::= ⟨adverbial modifier⟩ [⟨the group of adverbial modifiers⟩]

⟨adverbial modifier⟩ ::= ⟨simple adverbial modifier⟩ | ⟨adverbial clause⟩ | ⟨participial phrase⟩ | ⟨infinitive phrase⟩

⟨simple adverbial modifier⟩ ::= ⟨adverb⟩ | [⟨conjunction⟩] ⟨preposition⟩ ⟨basic noun phrase⟩ | ⟨basic noun phrase⟩ ⟨adverb⟩

The adverbial modifiers describe the place and the time of action, the purpose and the cause of action, the manner of subject and other functions.

In this article we have restricted ourselves to the description of only these constructions as the reasonably full description of formal grammar of English can require much more place.

To implement the syntax analysis one should describe the structure of sentences by means of the constructions of PROLOG. For these purposes *compound objects* and *functors* are used [9]. The structure of these compound objects is based on the Backus-Naur form.

## 3. Objects, actions and events

The main feature of our approach to the problem of natural language understanding is in the use of the *semantic database* to save the descriptions of objects, actions and events.

The semantic database includes the following predicates: *person, job, organization, project, action, event, message, place, tim, machine, thing, animal, relation, intelligence, cause, abstr, number.* The arguments (variables) of these predicates will be referred to as *fields*.

Consider, as an illustration, the predicates *person, action, and event*.

The predicate *person* describes human being. This predicate has the fields:

1. Code of *person*
2. Designation of person
3. Sex
4. First name

5. Last name
6. Additional data (other names, honorary title and degree)
7. Place of birth (code of *place*)
8. Nationality
9. Mother tongue
10. Other tongue (parallel with mother)
11. Place of residence (code of *place*)
12. Description of face
13. Description of nose
14. Description of constitution
15. Description of eyes
16. Description of hair
17. Date of birth (code of *tim*)
18. Stature
19. Temperament
20. Psychological type
21. Profession

The predicate *action* means a physical action. This predicate has the structure:

1. Code of *action*
2. Semantic type of action (PROPEL, MOVE, GO, TRANSFER, ...)
3. Sort of action (for real action - "real", for supposed action - "sup")
4. Negation of action ("not")
5. Tense ("pres", "past", "fut", "futpast", "mod")
6. Type of tense (for indefinite — "indef", for perfect — "perf", for continuous — "con", for perfect continuous — "perfcont", for indefinite passive — "indpassiv", for perfect passive — "perfpassiv")
7. Adverb used with verb
8. Verb describing action (in infinitive)
9. Subject of action (code of *person, machine, organization, animal*)
10. Object of influence (code of *person, machine, thing, organization*)
11. Object that the action is directed from (code of *person, organization*)
12. Object that the action is directed to (code of *person, organization*)
13. Result state of action

14. Start of action (code of *tim*)
15. Starting location of action (code of *place*)
16. Final location of action (code of *place*)
17. Tool or way of action (code of *machine, thing*)
18. Purpose of action (code of predicate describing this purpose)

Semantic classification of action is founded on fundamental investigation in [7] (with some additions and modifications).

The predicate *event* presents the change of state of object (*person, machine, thing, animal*):

1. Code of *event*
2. Sort of event (for real event — "real", for supposed event — "sup")
3. Subject changing state (code of *person, machine, thing*)
4. Designation of event
5. Scale
6. Time of event (code of *tim*)
7. Starting state (according to scale in use)
8. Result state (according to scale in use)
9. Location of event (code of *place*)
10. Way (tool) providing change of state
11. Object of influence
12. Tense ("pres", "past", "fut", "futpast", "mod")

The *scale* describing the state takes the value:

- for *health* — from -100 (death) to +100 (the best state of health);

- for *hunger* — from -100 (to die of hunger) to +100 (to be fed up);

and so on.

One can use physical scales too. For instance, *temperature* is estimated with the help of Celsius scale. Scales have been considered more widely in [7].

**4. Semantic analysis**

The structures formed as a result of syntax analysis are translated into the set of predicates of the semantic database. The following sequence of operations is used for the semantic analysis of simple declarative sentence:

1. The analysis of adverbial modifier at the beginning of sentence.
2. The analysis of subject.
3. The analysis of the group of objects.
4. The analysis of the group of adverbial modifiers.
5. The analysis of predicate.

*4.1. The analysis of subject*

The analysis of the subject is executed using the semantics of basic noun phrase and determinative construction for basic noun phrase (attributive clause, infinitive and participial phrase). The simple basic noun phrase is transformed to the predicate of type: *person, animal, organization, project, thing, machine, place, tim, abstr, disease*. To form these predicates one should use the following rules:

1. Belonging to the certain type of predicates is determined with the semantic code of the last noun of phrase (from the *dictionary of paradigms*).
2. If the attribute for the last noun is a event, then the predicate *event* is formed as well.

The attributive clause is the statement that is viewed according to the rules of the analysis of simple declarative sentence. The semantics of participial (infinitive) phrase is reduced to the analysis of participle (infinitive) and construction controlled with this participle (infinitive).

*4.2. The analysis of predicate*

The analysis of predicate is implemented for the predicate of main sentence as well as for the predicates of attributive, object, and adverbial clause. This analysis is realized for the verbal form of infinitive and participial phrase too.

The following predicates are formed as a result of the semantic analysis: *action* — for physical actions, *message* — for the transmission of information, *intelligence* — for feelings and thoughts, *job* — for long goal-seeking occupations, *event* — for events.

The choice of predicate is based on the indication of the semantic type in the *dictionary of verbs*.

Forming of semantic predicates describing actions and events is reduced to the determination of factors that are typical for these actions and events:

- the subject of action;
- the objects that take part in transmission of information or relation;
- the objects that the action is directed at;
- the location and time of action;
- the purpose (result) and method (tool) of action;

The use of these factors permits to overcome the problem of ambiguous expressions by means of choice of semantic variant corresponding to one of the descriptions of the given verb in the *dictionary of verbs* .

*4.3. The example*

Consider the sentence: *Mister Brown was a mate on a ship fifteen years ago*. The system LEIBNIZ forms the following set of predicates of the semantic database:

person(601302,"","m","Brown","","","","","","","","","","","","","","","","","","mate")
person(792287,"mate","","","","","","","","","","","","","","","","","","","mate")
machine(642100,"","ship","","","","",0,0,0)
tim(115429,"","years","","","","","","","","","","")
action(940765,"IS","","","past","indef","","be",601302,792287,0 , 0, 0, 115429, 0, 0, 0, 0, 0, "main", 527282,1)
cause(779419,"ago","",582914,"",115429)
number(582914,"fifteen","","",115429)
cadr(527282,473749,0,0,1)

## 5. The knowledge base

To form inquiry answers it is necessary to have the dictionary containing the explanations of words. This dictionary is realized in LEIBNIZ as the *knowledge base*. The knowledge base consists of the predicates *tperson, taction, torganization,...* that have precisely the same structure as the appropriate predicates of semantic database (*person, action, organization,...*). Building the knowledge base is founded on the algorithms of syntax and semantic analysis considered in the sections 4. The following semantic designations are used in the knowledge base:

- to point to the type of objects — the names of the predicates of database (*person, organization, place,...*) or the words describing the semantic class (*people, plant,...*)

- to describe the actions — the verbs whose sense is predetermined in system (*be, go, move, message, transfer, ...*)

Thus, there is the set of words whose sense is implied in LEIBNIZ and realized by appropriate algorithms. All the other words are explained by means of the knowledge base. These explanations are conceptually the semantic (verbal) definitions of words.

The knowledge base consists of *articles (frames)*. To form the article the phrase is entered:

*frame* is ⟨description⟩

and then the content of the article is described.

5.1. The article of noun

To form the article of noun (for example, *doctor*) the phrase of this kind is at first entered: *frame is doctor*.

Then the functions of this noun are indicated:

*doctor examines a person*
*doctor determines a disease*
*doctor prescribes a medicine*

The structure of object (for example, for the noun *car*) is formed by means of the phrase :

*A car consists of chassis, engine,...*

5.2. The article of verb

To form the article of verb (for example, *to go on*) the phrase of type *frame is go on* is entered. To describe the different values of verb in addition the group of noun can be used:

*frame is shoot from a gun*
*frame is shoot a person*

Then the semantic definitions and the descriptions of particle actions are entered. For example, after the phrase *frame is learn* there will be the descriptions:

*to do exercises*
*to answer a teacher*
*to visit a lesson in class*
*to study a textbook*

After semantic definitions the motives and the causes of action can be described. For example, after the phrase *frame is shoot a person* there will be the expressions:

*to kill the person by gun*
*to get money from the person as the subject is criminal*
*to pay off the person as this person outrages the subject*
*to annihilate the person as this person is the enemy of the subject*

5.3. The article of operation

The operation is a sequence of actions to execute the certain purpose. The algorithm of operation is based on principles of production systems [3]. All conscious human activity includes the set of different operations planed beforehand and modified in the course of realization.

The description of operation can involve several alternatives. Each alternative is entered with separate article. For example, for the purpose *to rob the organization* one can indicate the following description of the first alternative:

*frame is rob the organization*
*alternative 1 ; to go to organization*
*alternative 1 ; to come in at labor time*
*alternative 1 ; to neutralize a personal*
*alternative 1 ; to open safes using tools*
*alternative 1 ; to take moneys*
*alternative 1 ; to come out from organization*

The sequence of actions described above consists of stages. Each stage can contain action, object of influence, motive and cause, condition or other important circumstance, way (tool) of action.

## 6. Conclusion

By recognizing that natural language understanding consists in a thorough insight into essence of statements of this language, two basic approaches can apply (from the methodological point of view) to the problem of computer understanding: *psychological* and *ontological*.

The psychological approach models the mechanisms and the memory structures of verbal behavior, which have been formed in the course of human development. Human language understanding is based on collective experience (fundamental world knowledge) and individual experience as well as on emotions and feelings connected with this individual experience [5]. It means that it is necessary to model emotions and feelings as well as processes of learning and forming of associative relations.

For the knowledge domain describing the social behavior the realization of psychological approach presents a considerable difficulty at the moment. Therefore, this approach is best suited to create a robot executing operations in complex circumstances. This robot must orient well in an environment by means of the outer organs of vision, hearing or other ways using the algorithms of recognition [6].

The ontological approach proceeds from the assumption that natural language maps the structure of the outside world, which consists of objects implementing different actions or changing their states. Computer understanding is provided at the cost of appropriate information structures containing the descriptions of objects (people, things, machines, organizations), actions, and events for these objects. Furthermore, the explanatory dictionary is used for the definition of words semantics.

In this case computer understanding differs from human as emotions and associations for perception are absent. This approach is more suited for the knowledge domain describing the social behavior as it enables to find the valid answers for many questions concerning social facts saved in computer memory. Thus, the proposed class of questions realizes the partial Turing's test.

This work contains the ontological approach to understanding problem. Facts and objects from a sentence map into the semantic database, and to form inquiry answer the inference engine is used. Therefore, this approach is named (from the point of view of computer technology) *object-oriented* as well.

There is a class of questions concerning facts and objects that the com-puter system finds right answers for. To extend the class of questions it should be modified information structures and algorithms of processing. In future more advanced expert and information systems for the social application can be built up with the help of these understanding algorithms.